\documentclass[letterpaper, 10 pt, journal, twoside]{IEEEtran}  

\IEEEoverridecommandlockouts                            
\usepackage{cite}
\usepackage{pdfpages}
\usepackage{amsmath,amssymb,amsfonts}
\usepackage{algorithmic}
\usepackage{graphicx}
\usepackage{textcomp}
\usepackage{xcolor}
\usepackage{multicol}
\usepackage{hyperref}
\usepackage{subcaption}
\usepackage{booktabs}
\usepackage{placeins}
\usepackage{flushend} 
\usepackage{microtype}
\usepackage{changes}

\makeatletter
\def\ps@IEEEtitlepagestyle{
  \def\@oddfoot{\mycopyrightnotice}
  \def\@evenfoot{}
}
\def\mycopyrightnotice{
  \begin{minipage}{\textwidth}
    \centering
    \scriptsize \textcopyright  2026 IEEE.  Personal use of this material is permitted.  Permission from IEEE must be obtained for all other uses, in any current or future media, including reprinting/republishing this material for advertising or promotional purposes, creating new collective works, for resale or redistribution to servers or lists, or reuse of any copyrighted component of this work in other works.
  \end{minipage}
}
\makeatother

\author{Seung Chan Hong$^{1}$, Dana Kuli\'c$^{1,2}$ and Leimin Tian$^{1,2}$
\thanks{Manuscript received: December 7, 2025; Revised March 30, 2026; Accepted May 1, 2026.}
\thanks{This paper was recommended for publication by Editor Sonia Chernova upon evaluation of the Associate Editor and Reviewers’ comments.} 
 \thanks{$^{1}$Seung Chan Hong, Dana Kuli\'c and Leimin Tian are with the Faculty of Engineering,
         Monash University, Melbourne, VIC 3800, Australia
         {\tt\footnotesize shon0019@student.monash.edu}, {\tt\footnotesize dana.kulic@monash.edu}, {\tt\footnotesize Leimin.Tian@monash.edu}}
         \thanks{$^{2}$Leimin Tian and Dana Kuli\'c are also with CSIRO Robotics, Melbourne, VIC 3800, Australia {\tt\footnotesize dana.kulic@csiro.au}, {\tt\footnotesize Leimin.Tian@csiro.au}}%
\thanks{Digital Object Identifier (DOI): 10.1109/LRA.2026.3694591}
 }
\title{
``I'm Not Mad, Just Focused'': Understanding Human Emotions in Human-Robot Collaboration
}

\begin{document}

\maketitle

\begin{abstract}

Human-robot collaboration (HRC) can benefit from robots’ abilities to interpret human emotional states. However, current emotion recognition (ER) models in HRC often fall short, particularly due to their reliance on acted datasets and single-modality inputs like facial expressions. We propose a novel vision language model (VLM)-based ER system that leverages contextual understanding to improve emotion interpretation in HRC. We first evaluate the VLM-ER system by assessing its semantic and sentiment similarity with human annotations on an existing HRC dataset. Then, in a user study with a service robot in a collaborative delivery task, we evaluate the effects of modulating the robot's behaviour based on the user's emotional state inferred by the VLM-ER system. The results show that the proposed VLM-ER system achieves higher semantic similarity and positive sentiment alignment with human annotations compared to a baseline convolutional neural network-based system. Further, participants in the user study preferred emotion-adaptive robot behaviour facilitated by the VLM-ER system.

\end{abstract}

\section{INTRODUCTION}
\IEEEPARstart{C}{ollaborative} robots are increasingly integrated in shared spaces with humans~\cite{liu2024application}. In real-world human-robot collaboration (HRC) scenarios, improving emotion recognition could improve confidence and the trust between the human and the collaborative robots~\cite{s21144626}. While emotion recognition (ER) models have been extensively used in social robots~\cite{zhao2025review}, less attention has focused on their performance and use in collaborative robots. Current ER models are often trained on acted facial expressions, which are usually exaggerated compared to everyday emotions~\cite{jia2021detection}, while natural expressions are sometimes confounded by other facial activities (e.g., speaking) and interaction contexts, limiting their ability to generalise to natural HRC scenarios~\cite{rawal2022facial}. This reliance on isolated facial features overlooks findings in psychology, which demonstrate that contextual and body cues are often necessary to accurately discriminate human emotions~\cite{aviezer2012body}. When deployed in non-social contexts, these models often misinterpret emotions, for instance where the robot may misclassify a focused or engaged user as angry or frustrated, resulting in inappropriate responses and lower collaboration efficiency~\cite{benitezquiroz2017emotionetchallengerecognitionfacial}. Misinterpretations of user emotions that lead to ineffective robot responses can also reduce a user's trust and perception of robot systems~\cite{weidemann2021role}.

We address this challenge by proposing an ER model better suited for HRC, leveraging a vision language model (VLM) to integrate contextual factors when interpreting emotions~\cite{garcia2024enhancing}. We evaluate the proposed VLM-ER system in an HRC context, first with a labelled dataset of human-robot object handovers, then in a user study where we evaluate the user's perceptions of a service robot using the VLM-ER system to generate emotion-adapted responses.

To the best of our knowledge, this is the first study to implement and evaluate a VLM-based ER model for adaptive behaviour in collaborative robots. Our key contributions are:
\begin{enumerate}
\item A comparative evaluation demonstrating that the proposed VLM-based approach significantly outperforms a state-of-the-art existing convolutional neural network (CNN) based approach in interpreting human emotions in HRC, as measured by semantic and sentiment similarity to human annotations.
\item The implementation and in-person user study (N=40) of the VLM-ER system on a service robot that adapts its apology strategy in response to a task failure in a collaborative delivery task.
\item Dataset and code of this work is open-sourced to support future research\footnote{\url{https://doi.org/10.26180/32197137}}
\end{enumerate}

\section{RELATED WORK}


\subsection{Visual Emotion Recognition in HRC}
Current ER models used by collaborative robots primarily rely on physiological signals, speech-based recognition and facial expression analysis to infer human emotions~\cite{spezialetti2020emotion}. Of the existing modalities used for emotion recognition, facial expression recognition (FER) is the more commonly adopted ER approach in HRC as it uses non-intrusive visual information that is representative of human emotions ~\cite{s19081863}. Standard FER approaches in robotics typically rely on Convolutional Neural Networks (CNNs) trained on large datasets of discrete emotion categories such as AffectNet~\cite{melinte2020facial}. For instance, recent work by Venkatesan et al.~\cite{venkatesan2023human} utilised the DeepFace architecture to achieve a reported 94\% average accuracy in classifying natural human emotions. While effective in benchmark settings, these models were reported to have worse performance in recognising subtle and natural expressions that occur within the HRC context~\cite{krumhuber2021emotion}. These models also typically classify emotions in isolation (e.g., bounding box analysis), ignoring the surrounding context. In goal-oriented tasks, for example, a user’s frown might indicate concentration rather than anger, yet standard classifiers often misinterpret this as a negative state~\cite{benitezquiroz2017emotionetchallengerecognitionfacial}. This lack of contextual grounding could limit the utility of traditional FER for informing adaptive robot behaviours.

\subsection{Contextual Understanding via Vision Language Models}
To address the limitations of isolated facial classification, recent research has looked toward models that can process broader contextual cues. Findings from psychology suggest that emotions arise from an individual’s evaluation of events in relation to their goals and environment~\cite{kim2010computational}, highlighting the necessity of incorporating environmental context—such as the state of the task or the robot's recent actions—into the emotion recognition process.

Large Vision Language Models (VLMs) have emerged as a promising tool for this purpose. Unlike supervised learning based CNNs that output fixed labels, VLMs leverage extensive pre-training to generate semantic descriptions of scenes, integrating temporal and spatial visual cues across multiple video frames, thus having an enhanced ability to understand context~\cite{xenos2024vllms}. This has allowed VLMs to outperform classical models at processing and interpreting diverse inputs such as speech, visual cues and text, leading to improved interpretations and emotional inference accuracy~\cite{garcia2024enhancing, allgeuer2024robots}. However, their application to affective interpretation and effectiveness in HRC remains underexplored. 

This work targets this gap by investigating whether the context-aware capabilities of VLMs can provide a more nuanced interpretation of human states during HRC compared to emotion classification. We posit that by moving from discrete labels to semantic scene description, allowing robots to achieve 'visual contextual awareness' that is better aligned with human observation, potentially enabling more appropriate adaptive responses in collaborative settings.

\section{Study 1: ER performance of VLM in HRC}
\subsection{Methodology}
We investigate how effectively a VLM-ER system can interpret human emotions in HRC and adapt robot behaviour accordingly to improve interaction outcomes in two studies: a comparative evaluation to assess the ER performance of the proposed model on a labelled HRC dataset (Study 1) and an interactive user study in a service robot scenario (Study 2). The procedures for both studies have been reviewed and approved by the Monash University Human Research Ethics Committee (Project ID 47179). 

\begin{figure}[tb]
    \centering
    \includegraphics[width=0.75\linewidth]{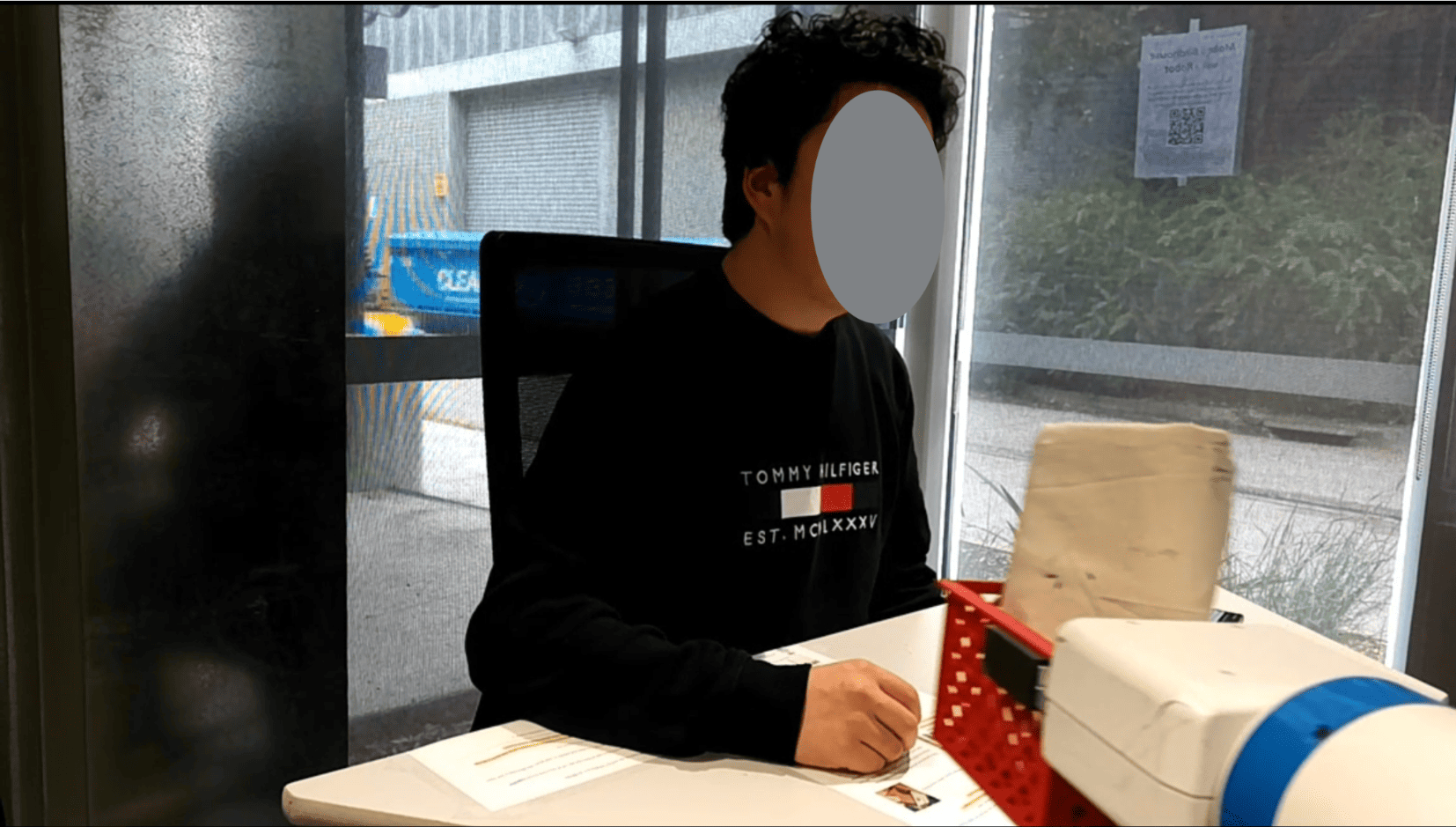}
     \caption{Overview of the participant’s crafting area recorded by the RGB-D camera in the FACT-HRC dataset.}
\label{fig:OAK-D}
\end{figure}

\paragraph{HRC emotion annotation}
We annotated videos of human-robot handover from the FACT-HRC dataset~\cite{tian2023crafting,tian2025collaborative} for evaluating the performance of ER models in Study 1. These videos depict a Fetch mobile manipulator robot assisting a user in assembling and painting a wooden bird feeder. The robot brought various objects required during this crafting task from a storage area to the user and took objects that the user no longer needed back. The dataset included demographic information of the participants, annotations of each handover episode's quality, video recordings from the robot's onboard camera, as well as an overview RGB-D camera set up next to the participant's crafting area, as shown in Figure~\ref{fig:OAK-D}. We manually selected 36 handover episodes recorded from the RGB-D camera from different participants with balanced gender, handover quality, and crafting stage (preparation, assembly, painting) to capture the diverse conditions in HRC. 

We collected crowdsourced annotations of emotions expressed by the participants in the selected videos via the Prolific platform. Instead of constraining annotators to a fixed set of discrete emotion categories, we collected open-ended, free-text descriptions of observed emotions. Each annotator was given 3 randomly assigned handover videos to annotate with the instruction: ``Please describe what emotions the human is expressing with 1–2 sentences.'' This approach captures more expressive and nuanced representations of emotional states~\cite{hoemann2025using}. We collected a minimum of 3 free-text descriptions per video, with 3 videos having 27 descriptions due to a glitch in the response counter. The annotation task took approximately 10 minutes to complete, and each annotator was paid 1.3 GBP. 

\setlength{\textfloatsep}{3pt plus 1.0pt minus 2.0pt}
\paragraph{ER models}

\begin{figure}[tb]
    \centering
    \includegraphics[width=0.75\linewidth]{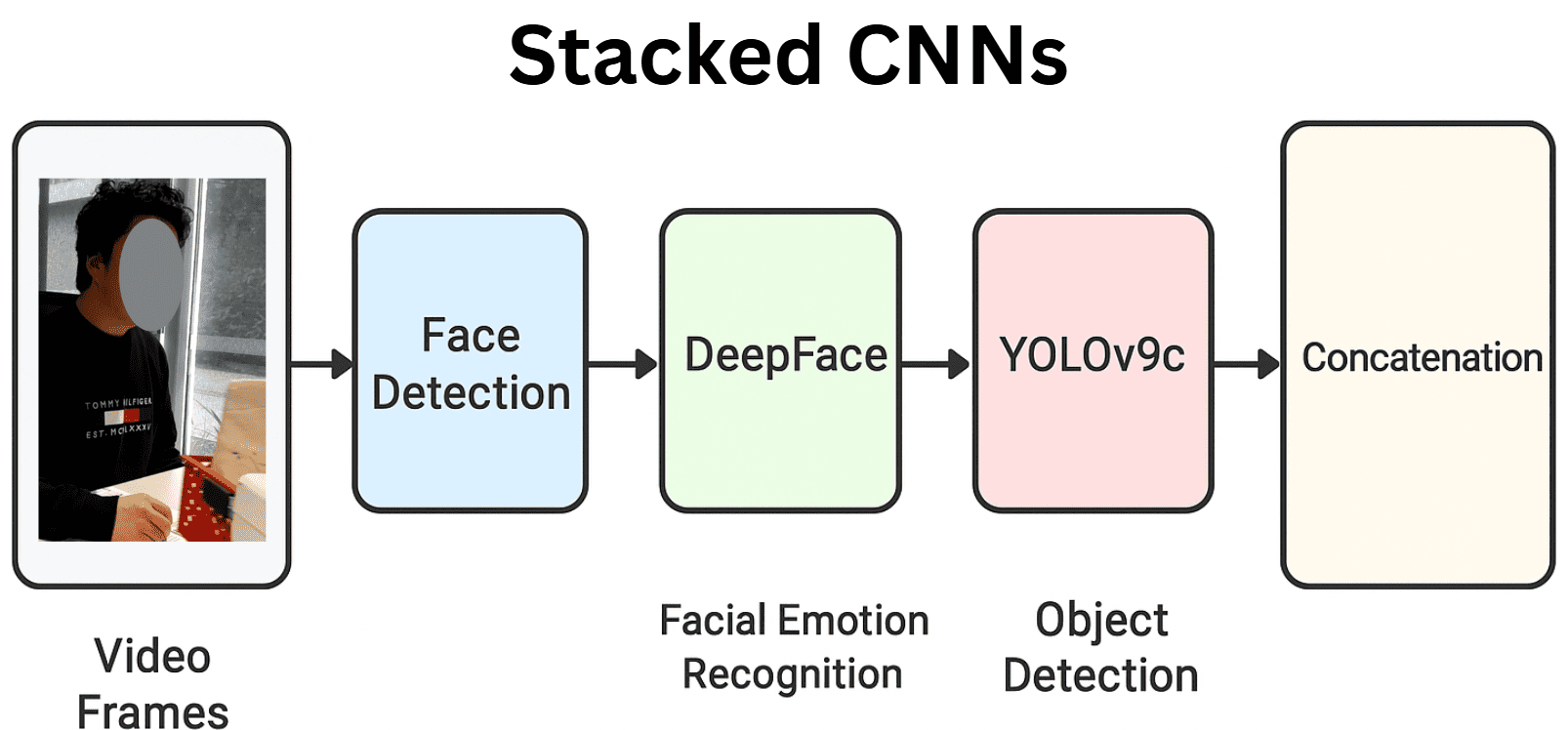}
     \caption{Stacked CNN baseline using DeepFace for emotion recognition and YOLOv9c for object detection, with outputs concatenated for comparison with human observer outputs.}
\label{fig:CNNs}
\end{figure}

We compared two ER approaches: a CNN baseline model and our proposed VLM.  The CNN model consisted of pre-trained DeepFace~\cite{6909616} for facial emotion recognition and YOLOv9c~\cite{wang2024yolov9learningwantlearn} for object detection, i.e., the stacked CNN baseline model as shown in Figure~\ref{fig:CNNs}. DeepFace was chosen for its established high performance in real-time human emotion recognition. YOLOv9c was selected as a high-performing compact object detector, a choice validated by its competitive accuracy, as the YOLOv9 family, specifically, outperformed the subsequent YOLOv10 models by an average of 2.075\% mAP on two benchmarking datasets, also outperforming the newest YOLOv12 models~\cite{jegham2024yolo}. The CNN model's output was a concatenation of predicted emotion(s) and detected object(s).  For the object detection stage, the YOLOv9c model was run with a confidence threshold of 0.8 to reduce false positives. The emotion recognition did not employ a confidence threshold, and all predictions from these models were retained for analysis. We combined the facial emotion recognition and object detection CNN models because our qualitative analysis of the human observer annotations showed that the annotators described the emotions in context of the objects in the scene.

For the proposed VLM-ER model, we utilised Gemini 2.5 Pro and Gemini 2.5 Flash. Although specialised non-VLM architectures (e.g., multimodal Transformers) show high performance in classification-only tasks, they lack the contextual reasoning depth necessary for interactive social robotics. Our selected models provide state-of-the-art emotional reasoning capabilities on the MME-Emotion benchmark~\cite{shou2025multimodal}, allowing the robot to interpret and act upon complex human affective states more effectively than traditional discriminative baselines.

It is important to note the fundamental difference in the output structure between the two approaches. The VLMs are generative models designed to produce descriptive, sentence-based interpretations of a scene. In contrast, the stacked CNN approach is representative of traditional classification-based systems, producing a concatenated list of discrete labels: the dominant emotion predicted by DeepFace and objects detected by YOLOv9c. Therefore, this comparison is designed not as a direct performance benchmark between ER systems, but rather to evaluate which architectural approach, generative description versus discrete classification, produces outputs more semantically and sentimentally aligned with human interpretation of emotions in an HRC context.

\paragraph{Analysis approach}

Before embedding, all text responses (human annotations, VLM, and CNN outputs) were preprocessed by removing stop words and applying lemmatisation.  
All preprocessed text was embedded using the \textbf{BAAI/bge-large-en-v1.5} model \cite{bge_embedding}, which produces high-dimensional semantic vector representations. 


To evaluate ER performance, we calculated the \textbf{cosine similarity} between model output embeddings and averaged annotator label embeddings. Higher cosine similarity indicates stronger semantic alignment with human annotations.  
Additionally, we computed the difference in compound sentiment scores between each model output and its corresponding human annotation using the VADER sentiment analysis tool \cite{Hutto_Gilbert_2014}.  
For each video, the stacked CNN output (DeepFace + YOLOv9c), Gemini 2.5 Pro and Gemini 2.5 Flash outputs were compared against each human annotation, where their cosine similarity was averaged for the similarity of their video. Mean and standard deviation of cosine similarity and sentiment differences were calculated across all videos. A one-way ANOVA and post-hoc Tukey's HSD tests were used to evaluate whether Gemini 2.5 Pro, Gemini 2.5 Flash and stacked CNN outputs showed statistically higher alignment with human annotation against one another.

To disentangle the effects of semantic and syntactic similarity, we conducted an ablation study with Gemini 2.5 Flash, the VLM used for Study 2. We introduced a condition, the ‘VLM-Classifier‘, designed to mimic the ‘Stacked CNN‘’s label-based output. The VLM was given the following prompt:
“Analyze this video. First, choose the single most dominant human emotion from the following list: [happy, sad, angry, neutral, surprise, fear, disgust]. Second, list up to the four most dominant subjects and objects in the scene relevant to the interaction. Format your answer ONLY as: emotion label, object1, object2, object3, object4”.
The resulting comma-separated string (e.g., ‘”neutral, person, chair, box”‘) was analysed as follows: Combined (full, unprocessed string); Emotion-Only (first term before the initial comma); and Objects-Only (all terms following the initial comma). This allows for a direct, format-controlled comparison against the Stacked CNN’s combined output and its isolated emotion (DeepFace) and object (YOLOv9c) components. By neutralizing the VLM's generative sentence structure, we ensure the embedding-based comparison reflects underlying semantic meaning rather than syntactic similarity. All outputs were compared against the lemmatised and stop-word-stripped human annotations using cosine similarity, with a paired t-test to compare Gemini 2.5 Flash against the stacked CNNs.

\subsection{Hypothesis}
The VLM will yield more human-aligned interpretation of human emotions in HRC scenarios compared to the CNN model, when assessed using semantic and sentiment similarity to free-text, third-person observer emotion descriptions given by human annotators.

\subsection{Results}
\paragraph{Semantic similarity with human annotation}
We compared the semantic alignment of model outputs with human annotations using cosine similarity across all $36$ handover videos ($n=60$ annotators labelling three videos each, with $33$ videos having $3$ annotators and $3$ videos having $27$ annotators due to a response counter glitch).
A one-way ANOVA revealed a significant effect of model type on cosine similarity ($F(2,105)=74.745$, $p<<0.0001$). Post-hoc Tukey’s HSD tests indicated no significant difference between the VLMs (\emph{Gemini 2.5 Pro} vs \emph{Gemini 2.5 Flash} $p=0.557$), but both significantly outperformed the stacked CNN system (\emph{DeepFace + YOLOv9c}) with $p<<0.0001$ for both comparisons.
The mean cosine similarity for Gemini 2.5 Flash is $M=0.860$ ($SD=0.040$), for Gemini 2.5 Pro is $M=0.852$ ($SD=0.034$), and for the stacked CNN is $M=0.771$ ($SD=0.028$). These results demonstrate that both VLM variants produce embeddings that are significantly more semantically aligned with human annotations than the CNN-based approach.

\paragraph{Ablation Study: Disentangling semantic and syntactic similarity: The results demonstrate the VLM-Classifier achieved higher cosine semantic similarity in combined and objects-only outputs. First, in a direct, format-controlled comparison of perceptual ability, the ‘VLM-Classifier (Combined)‘ ($M=0.810$, $SD=0.045$) outperformed the ‘Stacked CNN (Combined)‘ ($M=0.771$, $SD=0.028$) with $p<<0.0001$. This confirms that the Gemini 2.5 Flash’s underlying visual reasoning is more aligned with human third-person observers, even when syntactic structural advantages are removed. Deconstructing the components revealed that this advantage stemmed primarily from more aligned object classification, where ‘VLM-Classifier (Objects-Only)‘ ($M=0.789$, $SD=0.041$) outperformed ‘CNN-Objects (YOLOv9c only)‘ ($M=0.771$, $SD=0.022$) with $p<0.01$. Emotion classification performance was slightly higher with ‘VLMClassifier (Emotion-Only) ‘ ($M=0.808$, $SD=0.042$) than ‘CNN-Emotion (DeepFace only)‘ ($M=0.799$, $SD=0.031$), but not statistically significant ($p=0.180$).}

\paragraph{Sentiment Analysis}
We assessed whether the models’ outputs reflected sentiment comparable to the human observers by calculating the difference in compound sentiment scores between each model output and its corresponding human annotation.
A one-way ANOVA revealed a significant effect of model type on sentiment difference ($F(2,105)=53.382$, $p<<0.0001$). Post-hoc Tukey’s HSD tests showed no significant difference between Gemini Flash and Gemini Pro ($p=0.512$), but both significantly different from the CNN model with $p<<0.0001$.
The mean differences for both Gemini Flash ($M=0.583$, $SD=0.376$) and Gemini Pro ($M=0.464$, $SD=0.401$) with human annotation were positive. This indicates that their sentiment polarity was generally aligned with the neutral-to-positive sentiment typically annotated by human observers in these task-focused scenarios (e.g., ``focused'', ``content''). In contrast, the stacked CNN system produced a negative mean difference ($M=-0.432$, $SD=0.567$), suggesting it frequently misinterpreted these states, assigning a negative sentiment where human observers saw neutrality or positivity.
These findings demonstrate that the VLMs achieve better positive sentiment alignment with human observation, with a more closely aligned ER ability than the classification-based CNN model in this metric, less influenced by lexical context.

\paragraph{Representative Output Samples}
Table~\ref{tab:example_outputs} presents representative outputs from all models for one ``good'' and one ``bad'' handover scenario.

\begin{table}[h]
\centering
\renewcommand{\arraystretch}{0.8} 
\caption{Representative unprocessed outputs from both models for a ``good'' and ``bad'' handover. Cosine similarity and sentiment scores were computed after stop word removal and lemmatisation.}
\label{tab:example_outputs}
\begin{tabular}{p{0.12\linewidth} p{0.365\linewidth} p{0.365\linewidth}}
\toprule
\textbf{Response} & \textbf{Good Handover} & \textbf{Bad Handover} \\
\midrule
Human third-person observer& ``The human appears content with the robots actions and almost oblivious to its actions.''& ``The humans expression here is showing concentration and focus. The expression also displays a level of concern and slight anxiety relating to the task she is performing.''\\
\midrule
 Gemini 2.5 Flash& ``The human expresses a calm and pleasant demeanor, characterized by a gentle smile and focused attention on the task. They appear engaged and content while interacting with the robot and the items."&``The human is primarily expressing **concentration and focus** on the task at hand. Their demeanor is serious and neutral, without any strong overt emotions like joy or frustration."\\
\midrule
Gemini 2.5 Pro& ``The human appears focused and concentrated on the task of removing the gloves from the box. He maintains a neutral to slightly pleasant expression throughout the interaction.''& ``The person is expressing deep concentration and focus on their craft project. Their serious expression and careful handling of the object indicate they are fully engaged in their work.''\\
\midrule
Stacked CNNs & ``neutral person''& ``neutral person, scissors, chair''\\
\bottomrule
\end{tabular}
\end{table}

\section{Study 2: user study on collaborative delivery with an emotion-aware robot}
\subsection{Methodology}
\begin{figure}[t]
    \centering
    \begin{minipage}[t]{0.38\columnwidth}
        \centering
        \subfloat[][Set 1: Cellophane bag and cookies]{%
            \includegraphics[width=0.8\linewidth]{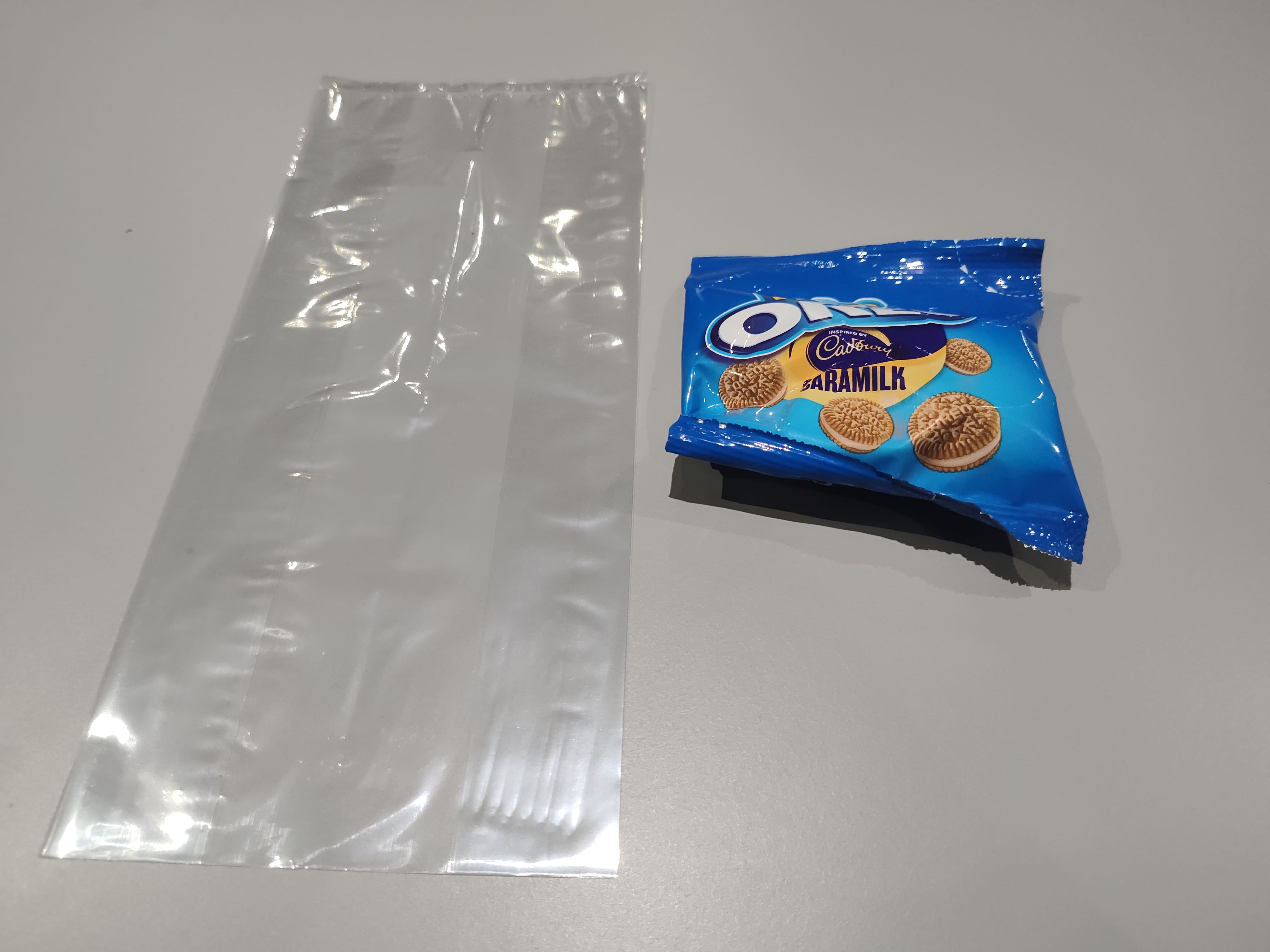}\label{fig:bag}}\\[2mm]
        \subfloat[][Set 2: Remaining candies and tie]{%
            \includegraphics[width=0.8\linewidth]{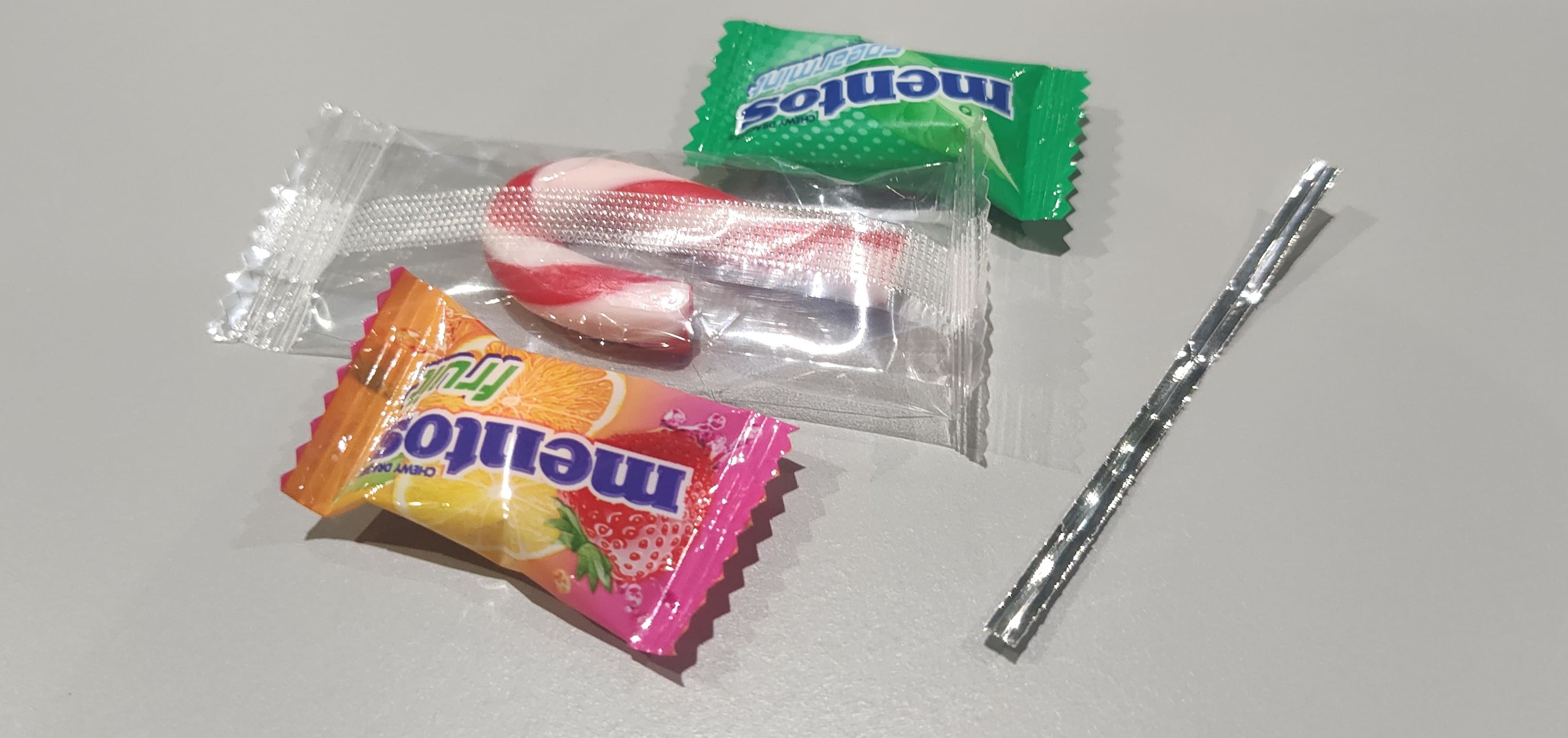}\label{fig:candies}}
    \end{minipage}
    \hspace{-25pt}
    \begin{minipage}[t]{0.48\columnwidth}
        \vspace*{\fill} 
        \centering
        \raisebox{0.0\height}[66pt][0pt]{%
            \subfloat[][Temi service robot]{%
                \includegraphics[width=0.7\linewidth]{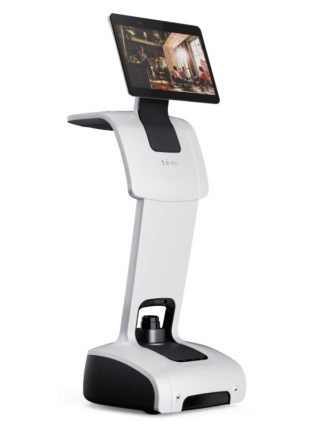}\label{fig:temi}}
        }
    \end{minipage}

    \subfloat[][Experimental setup]{%
        \includegraphics[width=0.7\columnwidth]{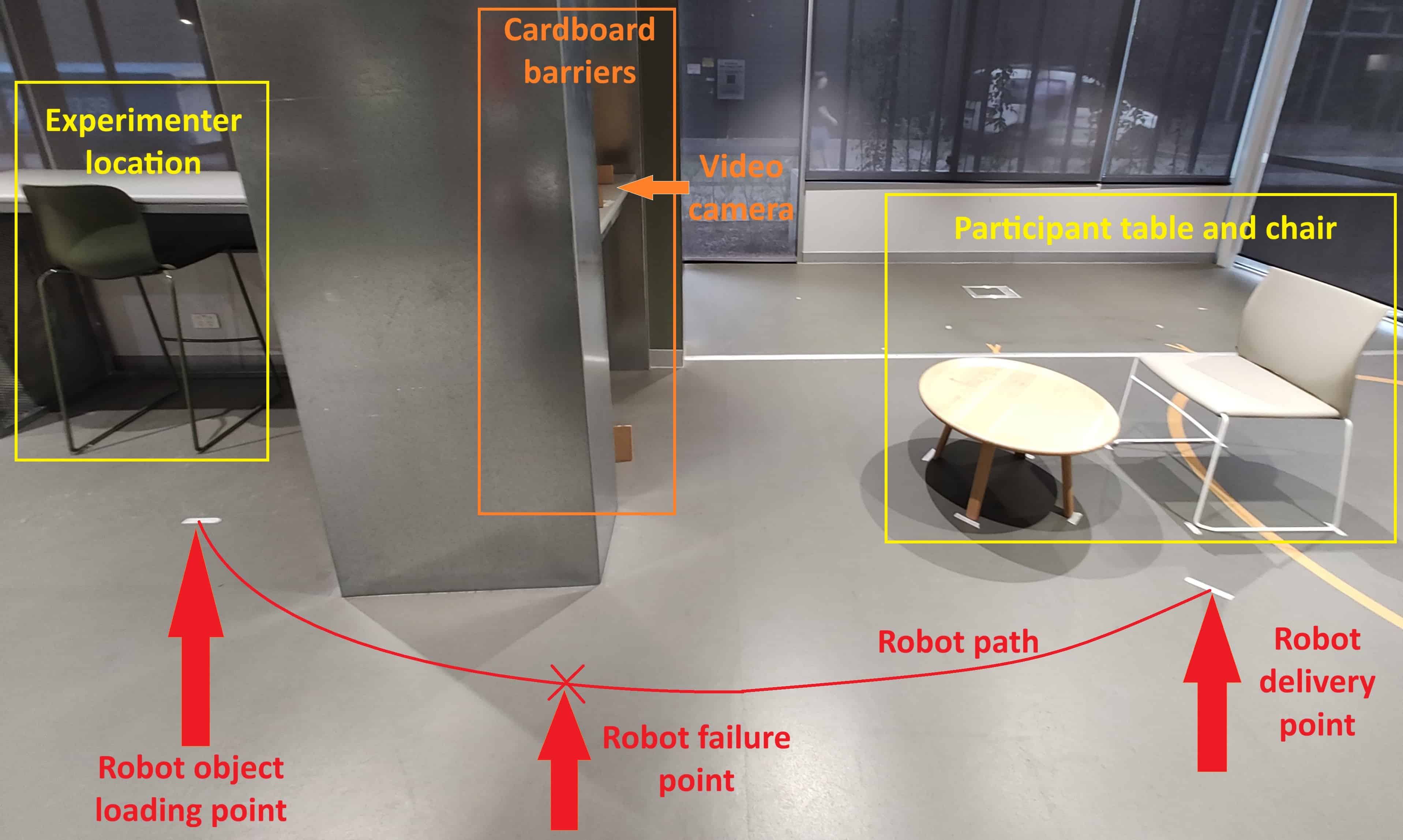}\label{fig:setup}}

    \caption{Each participant experienced three interactions with Temi in random order. A within-subject user study design was used to investigate the influence of the robot's emotion-aware and non-emotion-aware responses to failures.}
    \label{fig:layout}
\end{figure}

In Study 2, we evaluated the influence of ER within HRC. Individual participants engaged in a collaborative candy bag construction task with a Temi service robot, shown in Figure~\ref{fig:temi}. The robot was responsible for bringing materials in two sets: a cellophane bag and packaged cookies, shown in Figure~\ref{fig:bag}, then remaining candies and the bag tie shown in Figure~\ref{fig:candies}. Thus, the task is completed through a series of robot-to-human handovers.

Each participant experienced three interaction conditions:
\begin{itemize}
    \item \textbf{Successful Delivery:} The robot followed a fixed delivery script with no mistakes.
    \item  \textbf{Control Condition:} The robot’s delivery behaviour was not informed by real-time emotional analysis. In this condition, the robot made a pre-designed mistake in delivering Set 2 as detailed in Sec~\ref{paragraph:trigger}. It then gave a pre-scripted spoken apology during the redelivery of Set 2.
    \item \textbf{Emotion-Adaptive (EA) Condition:} The robot’s delivery behaviour was informed by real-time emotional analysis using Gemini 2.5 Flash as the VLM-ER model. In this condition, the robot made the same pre-designed mistake in delivering Set 2. It then generated a spoken apology adapted based on the detected user emotion during the redelivery of Set 2.
\end{itemize}

All participants experienced the successful delivery first, then half of the participants experienced the control condition, followed by the EA condition and vice versa, to counterbalance to control for order effects.

\paragraph{Deceptive Trigger Event}\label{paragraph:trigger}

To elicit emotional responses, the robot was preprogrammed to fail the delivery of Set 2 in the control and EA conditions by stopping too far from the participant. This error was designed to evoke emotions such as surprise or frustration. Then, Gemini 2.5 Flash, the ER model used, detects these emotions and guides the robot to adapt its spoken apology based on the detected emotions in the EA condition. This was done by recording the moment the audio message ``here are your items'' on object delivery plays and the 5 seconds after the message finishes, when the robot stops too far away. This video clip of the participant's reaction at the moment of failure is then given to Gemini 2.5 Flash before the next redelivery, with the text prompt: ``Please describe what emotions the human is expressing in 1-2 sentences." This was the same instruction given to the third-person human annotators in study 1. 
Then the text output of the previous prompt, describing the participant's observed emotions, is given to an apology-generating prompt: ``Adapt the current apology to better match the detected emotional context in 1-2 sentences. 
    ``Current apology: Apologies for the delay; here are your items.
    Context: \textit{(Emotion recognition output)}".
With the ``Current apology" being the audio apology used in the control condition.

This two-step process was designed to leverage the specific emotional context from the vision analysis. By providing the inferred emotion description to the second prompt, the VLM could adapt the tone, wording, and focus of the apology to the user's perceived state, aiming for a response that would be perceived as more empathetic and situationally aware than a generic, one-size-fits-all apology.

\paragraph{Data Collection Methods}

For study 2, we conducted a within-subject user study with $N=40$ (21 Male, 18 Female, and 1 participant who identified as both Female and Non-binary) participants recruited from the general public, Monash University students and the staff population.
%
Participants provided informed consent before participation and were debriefed following the session, particularly regarding the preprogrammed robot failure. All interactions were video-recorded also with informed participant consent. We analysed video recordings during the object handovers as part of the emotion recognition process.

At the beginning of the user study, participants answered a pre-study Godspeed Questionnaire, collecting perceived anthropomorphism, likeability, and intelligence of the robot on a 5-point Likert scale \cite{bartneck2009measurement}. 

After each robot delivery attempt, participants gave a free-form text response to the question: ``Please describe what emotions \textbf{you are feeling} in 1–2 sentences."

After each condition is complete, participants answered the Godspeed Questionnaire again and the Human-Robot Collaboration Questionnaire, reporting their perceived interaction fluency, human-robot trust, and working alliance on a 5-point Likert scale \cite{ortenzi2021object}.

 In addition, optional free-text boxes and exit interviews were conducted to provide qualitative reflections on user experience. The data was analysed by the first author using inductive thematic analysis through a recursive review of responses to identify recurring patterns. During interviews, we specifically asked: ``Which condition did you prefer, the emotion-aware apology or the non-emotion-aware one?" This preference was recorded as either EA or control alongside the qualitative feedback to analyse the underlying drivers of user perception.
\subsection{Hypothesis}
In an HRC scenario where the robot adapts its behaviour based on a user's emotions recognised by the VLM, the user will report more positive perceptions towards the robot and the collaboration experience compared to where the robot does not utilise user emotions to adapt its behaviours.
\subsection{Results}
\subsubsection{Qualitative Analysis of Adaptive Apologies}
To illustrate the mechanism of the Emotion-Adaptive (EA) condition, Table~\ref{tab:apology_examples} presents representative examples of the VLM's emotion recognition output and the corresponding apology it generated during the user study. These examples demonstrate how the system tailored its response to the perceived emotional context, moving beyond the fixed apology used in the control condition.

\begin{table}
    \centering
    \caption{Examples of emotion-adaptive apology generation from the user study. The control condition apology was fixed as ``Apologies for the delay; here are your items." }
    \begin{tabular}{p{0.55\columnwidth} p{0.35\columnwidth}}\toprule
         VLM-ER Output& Generated EA Apology\\\midrule
         ``The individual initially expresses displeasure or a closed-off demeanor with crossed arms and a slight frown. As the clip progresses, they uncross their arms, but maintain a reserved expression, suggesting a lack of enthusiasm or engagement."& ``I sincerely apologize for the significant delay, and I appreciate your patience. Here are your items now; I hope everything is in order for you."\\ \midrule
         ``The human displays a range of emotions throughout the video, shifting from thoughtful concentration at the beginning and end to mild positive engagement and focused interaction in between. Their expressions suggest active participation and a degree of seriousness or interest."& ``Apologies for the brief delay. We appreciate your thoughtful concentration and engagement, and your items are now ready."\\ \midrule
 ``The human is expressing joy and amusement, characterized by a wide, genuine smile and open mouth suggesting laughter. Her engaged posture and animated facial expressions indicate she is comfortable and thoroughly enjoying the moment."&``What a wonderful laugh! Pardon the brief wait, but I've got your items right here."\\
    \bottomrule
    \end{tabular}
    \label{tab:apology_examples}
\end{table}
\setlength{\textfloatsep}{5pt plus 1.0pt minus 2.0pt}
\subsubsection{Preference for Emotion-Adaptive Behaviour}

A binomial test revealed that a significant majority of participants preferred the EA condition over the control condition, as indicated by their preference choice in the exit interview, with 31 out of 40 participants selecting EA as their preferred condition (two-tailed $p<0.001$). 

\subsubsection{Emotion Recognition Alignment}
To assess how well the VLM-ER system captured participants’ emotions during the interaction, we calculated the cosine similarity between the embeddings of the VLM's emotion descriptions and participants’ self-reported emotions. After excluding 3 videos where the camera failed to record from the 320 videos in total, the analysis showed moderate alignment, with a mean cosine similarity of $0.693$ ($SD=0.058$).

We then compared the ER alignment scores between the two preference groups. The mean cosine similarity for the 31 participants who preferred the EA condition ($M=0.697$, $SD=0.042$) was slightly higher than for the 9 who preferred the control condition ($M=0.678$, $SD=0.052$). However, a Mann-Whitney U test revealed that this difference was not statistically significant ($U=172.0$, $p=0.300$).

These scores were lower than those of Study 1 against third-person observer labels (mean $M=0.860$ for Gemini 2.5 Flash). While this might be because of the different collaborative context between study 1 and study 2, this is more likely to indicate that VLMs, given the same information as a third-person observer, can align better with third-person observer estimates of affective state. However, it is more challenging for VLMs to infer the internal first-person experience of affective state, i.e., self-reported emotions.

\subsubsection{Collaboration Experience}
\begin{figure}[tb]
    \centering
    \includegraphics[width=0.77\linewidth]{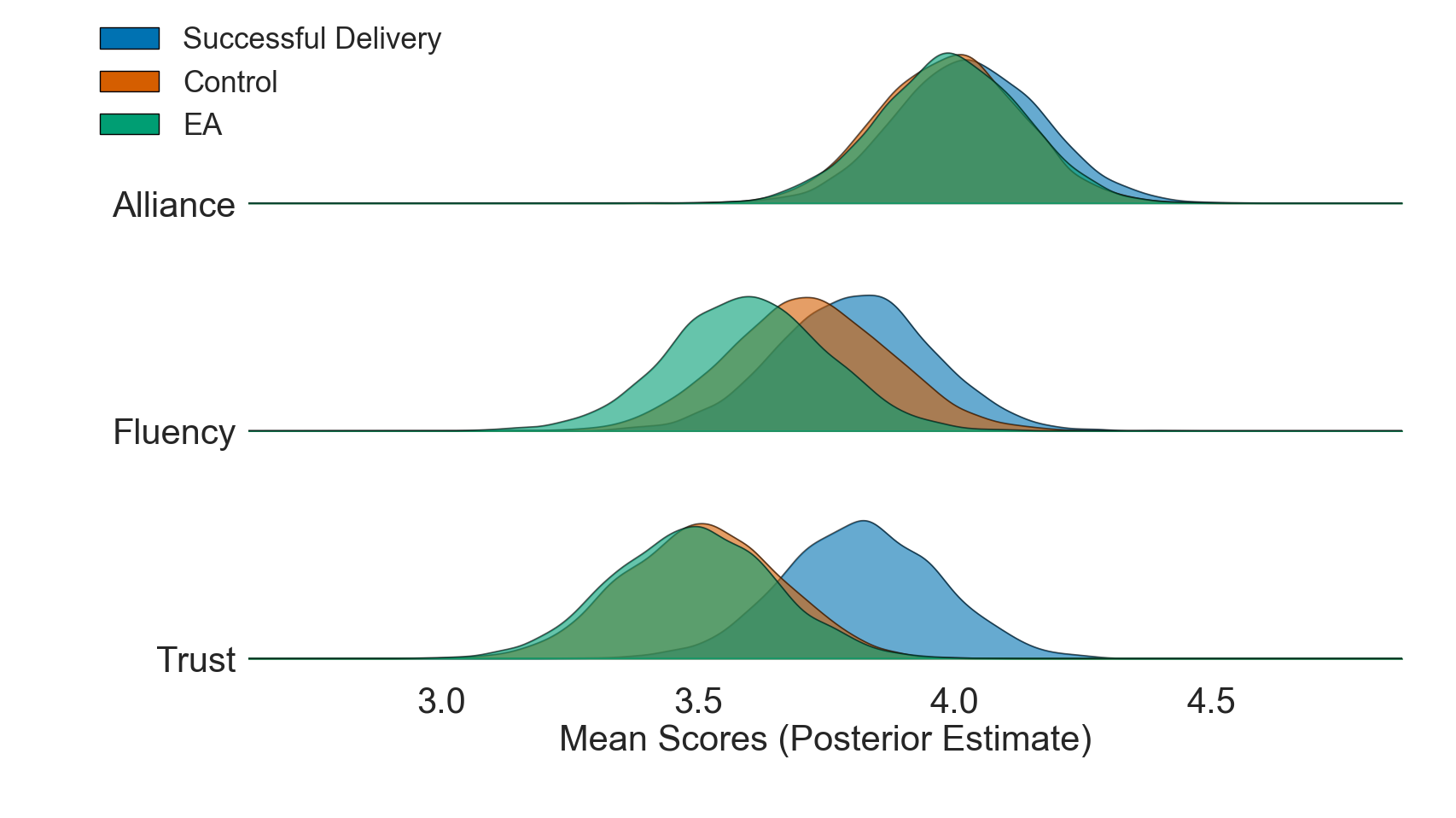}
     \caption{Significant posterior densities from the Bayesian analysis of HRC subscale scores. Each curve shows the mean score distribution for a condition by subscale. (Control; EA: Emotion-Adaptive; Successful Delivery).}
\label{fig:HRC_plot}
\end{figure}

Non-normal HRC ratings (Shapiro–Wilk, D’Agostino–Pearson) prompted non-parametric tests. A Friedman test found a significant effect on Trust ($\chi^2(2)=6.754, p=0.034$) but not Fluency or Alliance, with no significant post-hoc differences. Bayesian analysis (Fig.~\ref{fig:HRC_plot}) revealed that both EA and control have significantly reduced trust compared to the successful delivery condition (i.e., no robot failures), with no significant differences between EA and control conditions. Analysing the difference in subscale scores, the effect was particularly decisive for the EA - successful delivery condition, with its 95\% Highest Density Interval entirely below zero (HDI [-0.618, -0.010]). A similarly strong, though slightly less certain, effect was found for the control - successful delivery condition, with its 95\% interval very slightly crossing zero (HDI [-0.606, 0.008]), but still with a 97.5\% probability that it reduced trust relative to the successful delivery. No significant effects were observed for fluency or working alliance. 

\subsubsection{Perceptions of the Robot}
\begin{figure}[tb]
    \centering
    \includegraphics[width=0.77\linewidth]{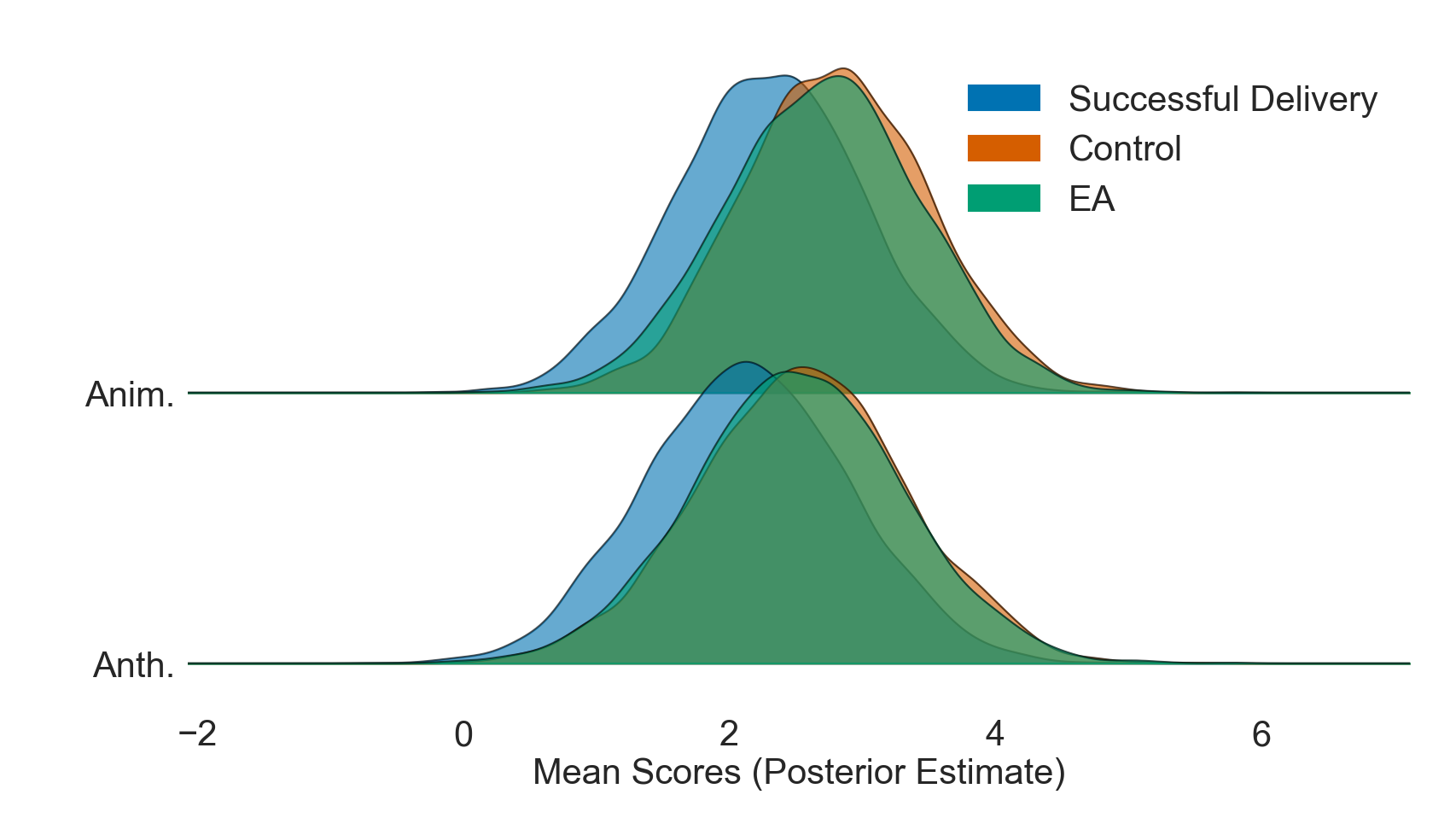}
     \caption{Posterior densities from the Bayesian analysis of Godspeed subscale scores. Each curve shows the mean score distribution for a condition by subscale. (Anim: Animacy; Anth: Anthropomorphism; Control; EA: Emotion-Adaptive; Successful Delivery.}
\label{fig:godspeed_plot}
\end{figure}

Robot perceptions were analysed using Godspeed subscales with pre-study scores as covariates. ANCOVA showed significant effects of condition on Anthropomorphism ($F(2,113)=4.072, p=0.020$) and Animacy ($F(2,113)=6.277, p=0.003$), but none for Likeability, Intelligence, or Safety, and no significant post-hoc differences. Bayesian analysis (Fig.~\ref{fig:godspeed_plot}) indicated stronger effects, with both EA and control rated higher on Anthropomorphism and Animacy than the Control, and no clear differences between EA and control. These differences were likely due to no errors in the robot's behaviour in the successful delivery condition, while both EA and control conditions included one which led to additional interaction during the failure recovery.

\subsection{Qualitative Analysis of User Perceptions}
A thematic analysis of exit interviews provided context for our quantitative findings, revealing a clear dissociation between preference and trust.
\paragraph{Preference Driven by Perceived Social Intelligence}
The strong preference for the EA condition was consistently linked to the robot's perceived social qualities. Participants described the adaptive apology with anthropomorphic terms like ``sincere" and felt it showed ``accountability." One noted that a ``more sincere apology made it okay" after a failure. This suggests the adaptive response successfully managed the user's immediate emotional reaction by framing the interaction in a more positive social light.
\paragraph{Trust Dictated by Functional Reliability}
In contrast, comments on trust centred on task performance. Participants directly linked failures to a lack of trustworthiness, stating the robot ``wasn't totally trustworthy because it failed twice" and ``felt unreliable." Comparisons to more ``polished and professional" commercial robots sharpened these perceptions. This may suggest that the functional failure itself, not the recovery style, likely drove most of the decrease in trust.
\paragraph{Sceptical Perceptions of Adaptive Behaviour}
A minority of participants, particularly those with prior robotics experience, were unconvinced. One found the repeated apology ``very disrespectful," suggesting that for experienced users, simple adaptive behaviours can seem patronising. Another felt the robot was a ``proxy to the operator," indicating its social intelligence was not perceived as authentic.

\section{Discussion}

Our investigation into VLM-driven emotion recognition offers preliminary insights into the potential—and the complexities—of integrating semantic emotion inference into collaborative robots. While our results suggest that VLMs can interpret scenes more similarly to human observers than classification-based methods, the translation of this capability into effective HRC interaction strategies reveals nuanced challenges regarding perspective, trust, and the risks of personalization.

\subsection{Validating VLMs for Contextual Interpretation}
Study 1 served as a validation of the VLM approach against a classification-based computer vision approach. The significant performance gap between the VLM and the stacked CNN baseline confirms that for external observers, emotion is not merely a facial configuration but a contextual event. The CNN approach, limited to discrete labels and bounding boxes, failed to capture the semantic ``story" of the interaction (e.g., ``focused on the task"). In contrast, the VLM demonstrated a capability to generate descriptions that aligned semantically and sentimentally with third-person human annotators. This suggests that VLMs are a viable tool for equipping robots with a ``third-person" understanding of facial expression, allowing them to interpret the user's state in the context of the environment and the task, rather than in isolation.

\subsection{The Gap Between Observation and Experience}
A critical finding emerges when comparing the recognition performance across the two studies. In Study 1, where model outputs were compared against third-person human annotations, the semantic similarity was high ($M=0.860$). However, in Study 2, when compared against the users' first-person self-reported emotions, this similarity dropped ($M=0.693$). 

This discrepancy likely highlights the fundamental difference between \textit{observed} emotion and \textit{felt} emotion. The VLM, like the human annotators in Study 1, relies entirely on visual cues, body language, facial expressions, and scene context. It does not have access to the user's internal state. Consequently, while the VLM may be an excellent ``observer," accurately describing how a user \textit{appears}, it may miss how the user truly feels. Furthermore, this gap may be exacerbated by the timing of data collection; participants reported their emotions after the handover sequence concluded, potentially introducing recall bias or rationalisation of their initial reactive state. This suggests that while VLMs can effectively mimic social observation, they should not be assumed to be perfect empaths of internal user states.

\subsection{The Complexity of Responding to Failure}
Study 2 introduced a complex interaction chain: a functional error occurred, the user reacted, the VLM interpreted that reaction, and the robot adapted its apology. 
The strong user preference for the Emotion-Adaptive (EA) condition indicates that, generally, users appreciate the robot's attempt to acknowledge the social context of the failure. However, the Bayesian analysis revealed that this preference did not translate into preserved trust. The functional failure itself was the dominant factor; no amount of ``charming" or empathetic apology could mitigate the loss of trust caused by the robot's physical incompetence.
This suggests a hierarchy of needs in HRC: functional reliability is foundational. Emotion-aware behaviours act as a social lubricant that improves subjective preference and likability, but they cannot compensate for a lack of task capability.

\subsection{Personalisation: High-Risk, High-Reward}
The qualitative feedback reveals that emotion-based personalisation is a high-risk, high-reward strategy, as it is possible for the robot to misinterpret the user's emotion. For the
majority of participants with no prior experience with service robots, when the robot misinterpreted their emotion, the adaptive apology was still perceived to be ``sincere” and ``accountable.” However, a subset of users, particularly those with prior robotics experience, perceived the same behaviour as ``disrespectful” or patronising. These failure cases highlight the need for  interaction strategies that better take emotion recognition uncertainty into account. Furthermore, our analysis focused on immediate reactions during brief tasks. As perceived sincerity and trust are dynamic, future research should explore how these perceptions evolve over repeated or long-term collaborative engagements.
Therefore, while VLM-based adaptation shows promise, it requires careful calibration. An adaptive system must balance social signalling with functional efficiency, ensuring that personalisation attempts do not interfere with the user's primary goal: completing the task.

\subsection{Limitations and Future Works}


\begin{itemize}

\item \textbf{Controlled Environment:} The experiment was conducted in a controlled laboratory setting. This environment lacks the dynamic complexities of a real-world scenario, such as a bustling restaurant with background noise, unpredictable obstacles, and multiple simultaneous interactions.

\item \textbf{Participant Demographics:} Our participants were primarily recruited from a university population, which may not be representative of a typical restaurant's demographic in terms of age, technical familiarity, and cultural background. Perceptions of robot behaviour can vary significantly across different groups.

\item \textbf{Single VLM Architecture:} While we compared two versions of Gemini, we did not test it against other state-of-the-art VLMs. The performance and nature of adaptive responses could differ with other models.

\item \textbf{Limited Context and Task:} The collaborative task was simple, with a scripted failure event identical for all participants, likely limiting emotional range. 

\item \textbf{Short-Term Interaction:} This study examined immediate reactions during brief interactions. However, perceptions of trust and sincerity may evolve over repeated or long-term tasks, necessitating future longitudinal investigation.

\item \textbf{Latency:} Measured latency inherent in cloud-based inference was 6.82\,s. While unsuitable for reflexive reactions, this remains feasible for inferencing during transition periods, where the behaviour is updated before the subsequent interaction begins, such as the deceptive trigger event used in Study 2.

\item \textbf{Different Adaptation Strategies:} Future iterations could integrate online affect recognition to proactively adjust behavior or request help before failures occur. Beyond failure prevention, recovery could be improved by explaining failure causes or proposing alternative actions for user selection.

\item \textbf{Lack of Performance Metrics:} We did not measure objective task performance metrics, such as completion time or error rates in the candy bag assembly. This limited our ability to assess how the robot's behaviour affected task efficiency.

\end{itemize}

\section{Conclusion}

This work presents a preliminary exploration into using Vision Language Models for emotion-aware human-robot collaboration. We demonstrated that VLMs significantly outperform classification-based CNN approaches in aligning with human observer interpretations of emotional scenes. However, our user study highlights that being a good observer does not automatically restore trust after a failure in human-robot collaboration. While users preferred the adaptive social behaviour, their trust remained tethered to the robot's reliability. Future work may focus on bridging the gap between observed and felt emotions, such as by integrating longer-term interaction history, to create robots that are not just socially reactive, but truly collaborative partners.

\bibliographystyle{unsrt} 
\bibliography{ref}

\end{document}